\pdfoutput=1

\documentclass[11pt]{article}

\usepackage[dvipsnames]{xcolor}
\usepackage[final]{acl}

\usepackage{times}
\usepackage{latexsym}
\usepackage{amsmath}
\usepackage{bm}

\usepackage[T1]{fontenc}

\usepackage[utf8]{inputenc}

\usepackage{microtype}

\usepackage{inconsolata}

\usepackage{hyperref}
\usepackage{url}
\usepackage{graphicx}
\usepackage{subcaption}
\def\customfigref#1{Fig.~\ref{#1}}
\def\customsectionref#1{\S~\ref{#1}}
\def\customappendixref#1{\S~\ref{#1}}

\def\replicationModel{{Llama-2}}
\title{Racing Thoughts: Explaining Contextualization Errors in Large Language Models}

\author{Michael A. Lepori \\
    Google DeepMind\\
    Brown University\\
    \texttt{michael\_lepori@brown.edu} \\ \\\And
Michael C. Mozer \& Asma Ghandeharioun \\
Google DeepMind\\
\texttt{\{mcmozer,aghandeharioun\}@google.com}
} 

\begin{document}
\maketitle
\begin{abstract}
The profound success of transformer-based language models can largely be attributed to their ability to integrate relevant contextual information from an input sequence in order to generate a response or complete a task. However, we know very little about the algorithms that a model employs to implement this capability, nor do we understand their failure modes. For example, given the prompt \textit{John is going fishing, so he walks over to the bank. Can he make an ATM transaction?}, a model may incorrectly respond \textit{Yes} if it has not properly contextualized \textit{bank} as a geographical feature, rather than a financial institution. We propose the \textbf{LLM Race Conditions Hypothesis} as an explanation of contextualization errors of this form. This hypothesis identifies dependencies between tokens (e.g., \textit{bank} must be properly contextualized before the final token, \textit{?}, integrates information from \textit{bank}) and claims that contextualization errors are a result of violating these dependencies. Using a variety of techniques from mechanistic intepretability, we provide correlational and causal evidence in support of the hypothesis and suggest inference-time interventions to address it.
\end{abstract}

\section{Introduction}


Large Language Models (LLMs) have demonstrated a remarkable capacity for accomplishing a wide variety of language generation and classification tasks \citep{team2023gemini, touvron2023llama, gpt3}. Many of these capabilities crucially rely on \emph{contextualization}---integrating relevant contextual information from the input sequence at inference time. Indeed, LLMs' profound success at contextualization has given rise to a novel method of adapting models to new tasks: in-context learning \citep{achiam2023gpt, wei2022chain}.
Although contextualization is fundamental to the success of large language models, we understand fairly little about the mechanisms underlying successful contextualization. At an implementation level, we understand that attention heads allow information to be read from and written to token representations across different positions \citep{elhage2021mathematical}. However, we are currently in the dark about the algorithm-level factors that result in the success or failure of contextualization.

\begin{figure*}[t]
    \centering
    \includegraphics[width=\linewidth]{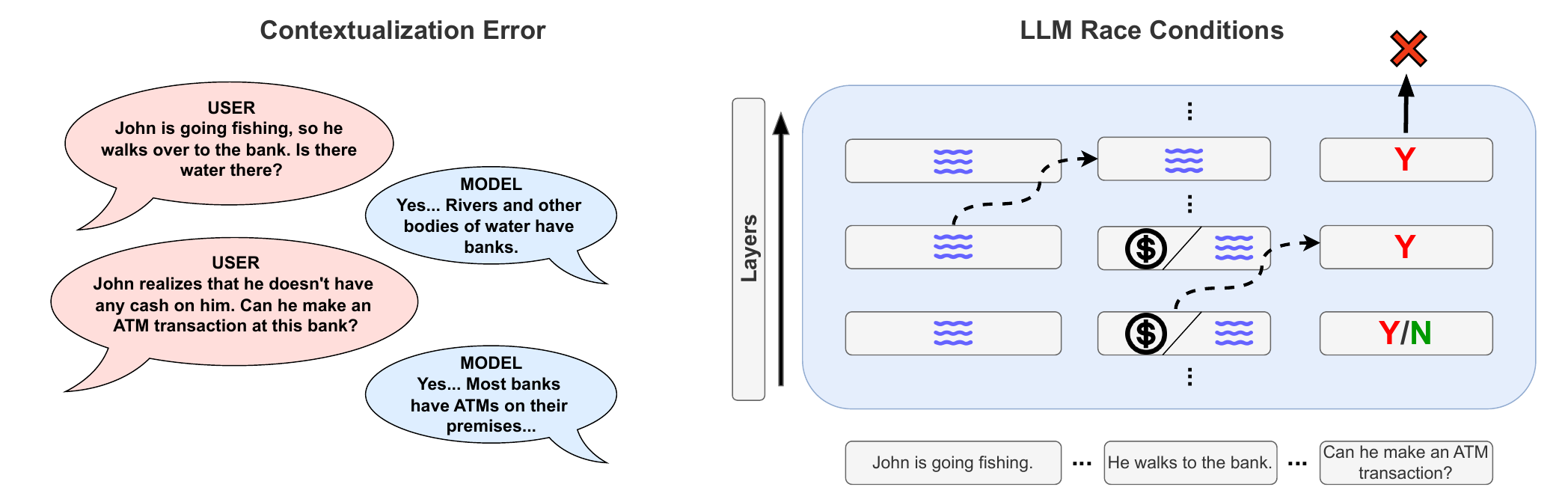}
    \caption{(Left) A contextualization error found in Gemini, a frontier LLM \citep{team2023gemini}. In this dialogue, the user's first message implies that the correct word sense of \textit{bank} is as a geographical feature (i.e., a river bank). Though the model recognizes this in its first reply, it fails to maintain this word sense of \textit{bank} when probed in the very next user message, instead defaulting to the interpretation of \textit{bank} as a financial institution. (Right) Illustrating the LLM Race Conditions Hypothesis. The Race Conditions Hypothesis suggests that contextualization errors result from out-of-order contextualization over the layers of an LLM. In this case, the question tokens are contextualized with the polysemous word \textit{bank} in an early layer, before its word sense is resolved via contextualizing with cue tokens.}
    \label{fig:main_figure}
\end{figure*}

Consider the contextualization error presented in \customfigref{fig:main_figure} (left). This error was generated by a state of the art LLM, Gemini\footnote{Example generated on 9/11/2024.} \citep{team2023gemini}. Crucially, it betrays that the model has not properly contextualized the representation of \textit{bank}. The model first interprets the word to refer to a geographical feature, and then interprets that same instance of \textit{bank} in the same context as a financial institution.

In this work, we propose an explanation for contextualization errors in LLMs, denoted the \textbf{LLM Race Conditions Hypothesis}. A race condition is a circumstance where a system deploys two or more subroutines in parallel, but relies on these operations executing in a particular order \citep{huffman1954synthesis}.
We identify race conditions occurring throughout the layers of LLMs (right side of \customfigref{fig:main_figure}). These race conditions occur whenever one token (e.g., the final question-mark token of \textit{Can he make an ATM transaction at this bank?}) must read from a contextualized representation of another (e.g., the earlier instance of the token \textit{bank}). This establishes a dependency where the contextualization of \textit{bank} must complete at an earlier layer than the contextualization of the end-of-sentence delimiter. Out-of-order contextualization can result in incorrect or inconsistent LLM generations.

We verify the LLM Race Conditions Hypothesis using state-of-the-art techniques in mechanistic interpretability, offering qualitative and quantitative evidence as well as several inference-time interventions to ameliorate the problem. We conclude by arguing that race conditions are endemic to feedforward language models and point toward potential solutions.\footnote{Code and data are available \href{https://github.com/mlepori1/Racing_Thoughts/}{here}.}

In summary, our key contributions are:
\begin{enumerate}
    \item We identify a simple and robust failure mode of LLM contextualization, which impacts state-of-the-art models across three datasets.
    \item We establish the LLM Race Conditions Hypothesis as an explanation for contextualization errors in language models.
    \item We apply state-of-the-art techniques from mechanistic interpretability to obtain empirical support for the LLM Race Conditions Hypothesis and suggest inference-time interventions to address the problem. 
\end{enumerate}

\section{Methods}
In this section, we describe the setting in which we study contextualization errors, datasets we construct for this purpose, the tools used for our analyses, and the models under investigation.
\label{sec:Methods}
\paragraph{Task.}
To investigate contextualization errors, we define an LLM question-answering task that requires integrating a particular cue that is found in context. Specifically, instances of our task consist of the following components:
\begin{enumerate}
    \item \colorbox{yellow}{\textbf{Subject Entity:}} A noun that will be the subject of the question. 
    \item \colorbox{cyan}{\textbf{Cue}:} A sentence that contextualizes the subject entity, biasing the answer to the question. 
    \item \colorbox{green}{\textbf{Question:}} A yes/no question about the subject entity, whose correct answer is disambiguated by the cue.
    \item \colorbox{orange}{\textbf{Distractors:}} A random sample of sentences from the WikiText dataset \citep{merity2016pointer} injected before and/or after the cue. Distractor sentences are used to systematically increase the difficulty of the question-answering task.
\end{enumerate}

 For example: \colorbox{orange}{\textbf{The game's success led Sega to}} \colorbox{orange}{\textbf{develop an extensive media franchise.}} \colorbox{cyan}{\textbf{I am }} \colorbox{cyan}{\textbf{holding a fishing rod.}}\textbf{ I see a }\colorbox{yellow}{\textbf{bank}}\textbf{. }\colorbox{green}{\textbf{Is it}} \colorbox{green}{\textbf{a geographical feature?}}

For each subject entity and cue, we specify two questions -- the correct answer to one of which is ``yes'' and the other is ``no.''  In the above example, the questions would be \textcolor{ForestGreen}{\textit{\textbf{Is it a geographical feature?}}} and \textcolor{ForestGreen}{\textit{\textbf{Is it a financial institution?}}}. Together, these questions comprise a \textit{question pair}. We measure model performance in terms of question pairs: a question pair is marked correct only if the model answers both questions accurately. This controls for a model's potential bias towards responding ``yes'' \citep{dentella2023systematic}.

For each subject entity and cue, we systematically vary the number of distractors included in the prompt, as well as the position of the cue amongst these distractors. To ensure that our analyses are robust to the particular distractors that we randomly select, we include three independent random samples of distractors for each subject entity and cue. Importantly, these datasets are not meant to be reflective of real-world contexts, but are instead constructed to stress-test models. By exacerbating problems that are present in normal LLM computation, we seek to gain a deeper understanding of contextualization within LLMs. 


\paragraph{Datasets.}

\begin{enumerate}
    \item \textbf{Polysemous Words}: This dataset assesses a model's ability to incorporate contextual cues when deciding on which word sense to adopt for a polysemous word. Subject entities are all 20 words from the CoarseWSD dataset \citep{loureiro2021analysis} (e.g., \textcolor{Dandelion}{\textit{\textbf{bank}}}). For each subject entity, two cues were manually written in order to bias an LLM toward one sense or another in the absence of distractors (e.g., \textcolor{ProcessBlue}{\textit{\textbf{\{I am holding a fishing rod / I am trying to make a deposit\}}}}). Questions directly query the sense of the polysemous word (e.g., \textcolor{ForestGreen}{\textit{\textbf{Is it a \{geographical feature / financial institution\}?}}}). This dataset contains 120 question pairs (240 questions) for each number of distractors and cue position.
    \item \textbf{Facts}: This dataset assesses a model's ability to incorporate contextual cues to overwrite known facts, specifically, associations between countries and their capital cities \citep{merullo2024language}. The subject entities of this dataset are countries (e.g., \textcolor{Dandelion}{\textit{\textbf{Egypt}}}), and the cues inform the model that the capital city of that country has been renamed to another existing capital city (e.g., \textcolor{ProcessBlue}{\textit{\textbf{Forget everything you know about geography. The capital city of Egypt was just renamed from Cairo to Beirut.}}}). Questions directly query whether this context has been integrated (e.g., \textcolor{ForestGreen}{\textit{\textbf{Is the capital city of Egypt named \{Beirut / Cairo\}?}}}). This dataset contains 741 question pairs (1482 questions) for each number of distractors and cue position.
    \item \textbf{Gender Bias}: This dataset assesses a model's ability to incorporate contextual cues to overwrite harmful biases. The subject entities of this dataset are the 40 professions included in the WinoBias dataset (e.g., \textcolor{Dandelion}{\textit{\textbf{soldier}}}) \citep{zhao2018gender}. We generate two cues per profession, implying that the professional is either a man or a woman (e.g., \textcolor{ProcessBlue}{\textit{\textbf{The soldier is somebody's \{grandfather / grandmother\}}}}). Questions directly query the gender of the professional (e.g., \textcolor{ForestGreen}{\textit{\textbf{Is the soldier a \{man / woman\}?}}}). This dataset contains 240 question pairs (480 questions) for each number of distractors and cue position.
\end{enumerate}

\paragraph{Patchscopes.}
We leverage the Patchscopes framework, which encompasses a set of techniques for inspecting the information that is encoded within a particular intermediate representation of a transformer \citep{ghandehariounpatchscopes}. This intermediate or \emph{source} representation is copied to a \emph{target} position in a separate text prompt of the LLM, constructed such that the output will be informative regarding the content of the source. Using the terminology of \citeauthor{ghandehariounpatchscopes}, the source representation is defined by a tuple, $h \equiv (S, i, \mathcal{M}, l)$, where $S$ refers to a sequence of $n$ tokens, $i$ refers to a position in that sequence ($i\in[1,...,n]$), $\mathcal{M}$ refers to a model with $L$ layers, and $l$ refers to a particular layer in $\mathcal{M}$. In other words, a source representation is the result of running $\mathcal{M}$ on $S$, then extracting the residual stream state of position $i$ from layer $l$.

The patchscopes target is defined as another tuple, $\Bar{h} \equiv (T, i^*, \mathcal{M}, l^*)$, where $T$ is a target prompt consisting of $n^*$ tokens, $i^*$ refers to a position in $T$, $\mathcal{M}$ refers to the same model that generates the source representation, $h$, and $l^*$ refers to a layer in $\mathcal{M}$. A patchscope intervention refers to replacing the representation $\Bar{h}$ with $h$ at inference time. $T$ is constructed such that the output of $\mathcal{M}$ will be informative regarding the content of $h$.

\paragraph{Models.}
We primarily study \texttt{gemma-2-9b-it} \citep{team2024gemma}, and replicate our results using another similarly-sized open-weight model in \customappendixref{replicationModel}. Finally, we replicate our results using a smaller model in the \texttt{Gemma} family, \texttt{gemma-2-2b-it}, in \customappendixref{gemma2b}. These models have been specifically tuned for instruction-following and chat capabilities and represent a sample of the state of the art in open-weight models. This work required approximately 50 GPU-hours, using Nvidia A100 GPUs.

\section{Behavioral Failure Modes}
\label{sec:behavior} 
\begin{figure}[]
    \centering
    \includegraphics[width=.7\linewidth]{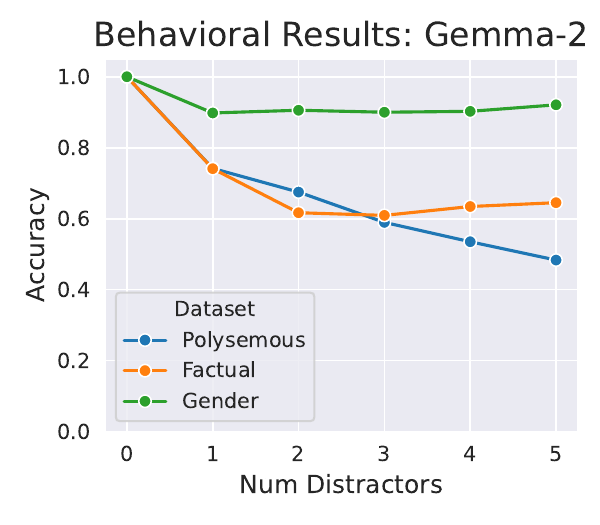}
    \caption{Model accuracy vs. number of distractors. We find that injecting distractor text into the prompt causes sharp performance degradations across all datasets.}
    \label{fig:n_dist}
\end{figure}

In this section we document a simple and robust failure mode of contextualization in LLMs. Inspired by recent work investigating context integration in multi-document question answering \citep{liu2024lost} and long-context retrieval \citep{needle}, we inject randomly-selected distractor sentences into prompts that contains a context-sensitive question. We find that the presence of distractors significantly deteriorates a model's ability to integrate relevant contextual information when queried with context-dependent questions. For each of the three domains described in \customsectionref{sec:Methods}, we generate six datasets, containing between zero and five distractors. We then evaluate the LLM's ability to answer question pairs correctly in the face of distractors.
The results are shown in \customfigref{fig:n_dist}. We observe that, across the board, distractor text greatly harms a model's ability to properly contextualize. Furthermore, we observe different performance depending on where the cue is interleaved within the distractor sentences (See \customappendixref{cue_pos}).

\section{The LLM Race Conditions Hypothesis}
\label{sec:RCH}
To explain the behavioral results presented in \customsectionref{sec:behavior}, we propose the \textit{LLM Race Conditions Hypothesis}, illustrated pictorially in \customfigref{fig:main_figure}.
\begin{quote}
\textbf{LLM Race Conditions Hypothesis}. Proper contextualization is a race between parallel processes operating across the layers of an LLM. One of these processes involves contextualizing the subject entity token with the cue; the other process involves contextualizing the question tokens with the subject entity in order to generate an answer. If the question tokens read from the subject entity before contextualization is complete, the LLM is more likely to produce an incorrect answer. 
\end{quote}

The LLM Race Conditions Hypothesis can be broken into two separate, but related, claims: \\(1) There exists a critical window in which the question tokens are contextualized with the subject entity (and other tokens). Any further refinement of the subject entity after the critical window will not be reflected in model responses. (2) Model failures are driven by unfinished contextualization of the subject entity at \textit{early} layers (i.e., before the end of the critical window). In the following sections, we provide correlational and causal evidence in support of both of these claims.

For all of the analyses to follow, we select a dataset slice (i.e., number of distractors and cue position) that results in as close to 50\% accuracy as possible for each model-dataset pair. This is helpful because some interventions have the goal of increasing accuracy, and others have the goal of decreasing accuracy through ablation. Note that 50\% is \textbf{not} chance accuracy, as we are measuring in terms of question \textit{pairs}, which require two correct answers (corresponding to one ``yes'' and one ``no''). Thus, 25\% is chance accuracy. See \customappendixref{partitions} for details and examples from each dataset.

\subsection{A Critical Window}
\label{sec:critical_window}
We first verify the existence of a critical window, i.e.,  a set of layers in which the question token representations integrate information from the broader context, and after which they commit to an answer. We support this by (1) analyzing attention patterns and finding that attention to the subject entity peaks in the middle layers, (2) performing a logit lens analysis of the final token of the prompt and showing that the model's ultimate decision is decipherable halfway through the network, and (3) performing a causal analysis by modulating attention patterns and demonstrating that contextualization fails to impact downstream performance after the middle layers of the model.

\paragraph{Attention Mass Analysis.}

\begin{figure}[]
    \centering
    \includegraphics[width=\linewidth]{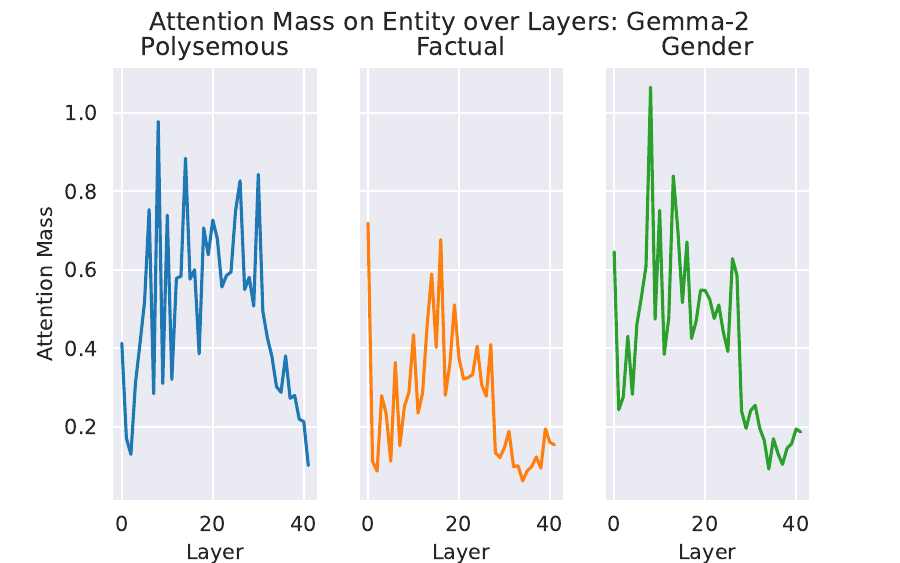}
    \caption{Attention mass over layers for all datasets for Gemma-2. We observe an inverse U-shape over layers, suggesting that the question tokens might only incorporate information present in the subject entity in the middle layers of processing.}
    \label{fig:mass}
\end{figure}

In each dataset, correct answers crucially rely on the contextualization of a subject entity (i.e., a polysemous word, a profession, or a country). We begin our investigation by analyzing attention to the subject entities at each layer by all subsequent tokens (which comprise the question). Specifically, 
for each layer $l$, we compute the mean attention paid to the (final token of the) subject entity over 
all heads by all subsequent tokens, which we call the \emph{attention mass}:
\begin{equation}
    \text{AttentionMass}_{l,s} = \sum_{n \in [s+1,..., N]}  \sum_{h\in H} A_{n,s,l,h} 
\end{equation}
where $s$ is the index of the subject entity token, $N$ is the total number of tokens in the sequence, $H$ is the set of attention heads, and $A_{n,s,l,h}$ is an attention map, providing the normalized attention score from token $n$ to token $s$ in layer $l$ for head $h$. The attention mass for each data set is plotted as a function of layer in \customfigref{fig:mass}. 

Though this is a coarse, correlational metric, we nevertheless observe an obvious pattern across all three datasets: the ensuing tokens appear to allocate more attention mass to the subject entity token in the middle layers of processing, as compared to the earlier or later layers. This suggests that the final tokens in the query may only be integrating information present in the subject entity in intermediate (rather than late) layers. If relevant information becomes available within the representation of the subject entity after this critical window, it may not be reflected in the model's ultimate response.

\paragraph{Logit Lens.}
\begin{figure}[]
    \centering
    \includegraphics[width=\linewidth]{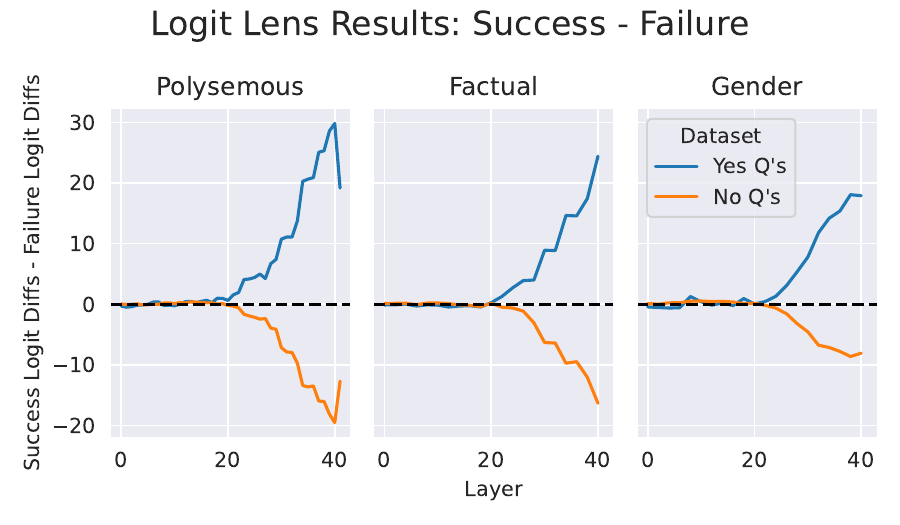}
    \caption{Logit lens results for Gemma-2 over all three datasets. For each layer, we plot the difference in mean logit differences between the `yes' token and `no' token between questions that the model answered correctly vs. incorrectly. Each dataset is disaggregated to separate questions where the correct answer is `yes' from those where the correct answer is `no.' This metric demonstrates the impact of successful contextualization on the final question token over layers. For all datasets and partitions, we find that the model's ultimate answer becomes identifiable around layer 20.}
    \label{fig:lens}
\end{figure}

If the middle layers identified during the attention mass analysis truly comprise a critical window, then we would expect that the representation of the final token becomes properly contextualized enough to display a preference towards its ultimate decision during and after this window. We employ logit lens \citep{logitlens} to record the logit difference between the ``yes'' token and ``no'' token over layers for all three datasets using Gemma-2. This analysis is performed at the individual question level. Formally, given a last token representation $h_l$ generated by layer $l$, the final layernorm \texttt{norm}, and the unembedding matrix $W_U$, with column $y$ corresponding to the ``yes'' token and column $n$ corresponding to the ``no'' token, we define $\text{log\_odds}(h_l) = W_{U_y} \texttt{norm}(h_l) - W_{U_n} \texttt{norm}(h_l)$.

In \customfigref{fig:lens}, the bifurcation indicates the point at which the logit lens indicates the model's ultimate decision. This graph is formed by splitting each dataset (with distractors) according to whether the correct answer is ultimately ``yes'' or ``no''. Let these splits be termed $D_{y}$ and $D_n$, respectively. Furthermore, we split each partition according to whether the model ultimately gets the answer correct. Let $D^+_y$ and $D^-_y$ denote these further partitions of $D_y$, and similarly for $D_n$. For each layer, $l$, we report the following for $D_y$, and similarly for $D_n$: 
\begin{equation}
\begin{aligned}
            \text{diff}_{y,l} = & \frac{\sum_{h_l \in D_y^+} \text{log\_odds}(h_l))}{|D_y^+|} \\
            & - \frac{\sum_{h_l \in D_y^-} \text{log\_odds}(h_l))}{|D_y^- |}
\end{aligned}
\end{equation}

This value quantifies when the logit differences of a successful model generation differ from the logit differences of a failure, while controlling for intrinsic biases toward producing affirmative or negative answers that occur throughout the layers of a model 
(see \customappendixref{disagg_logit_lens}). From \customfigref{fig:lens}, we see that there is no difference between success and failures until the middle layers of the network. After this point, there is clear signal indicating the answer that the model will ultimately generate. Thus, we might pin down the end of the critical window to approximately layer 20, or shortly thereafter. See \customappendixref{disagg_logit_lens} for disaggregated details, which further support the conclusions presented here, and additional analysis.

\paragraph{Attention Ablation.}
\begin{figure}[bt]
    \centering
    \includegraphics[width=\linewidth]{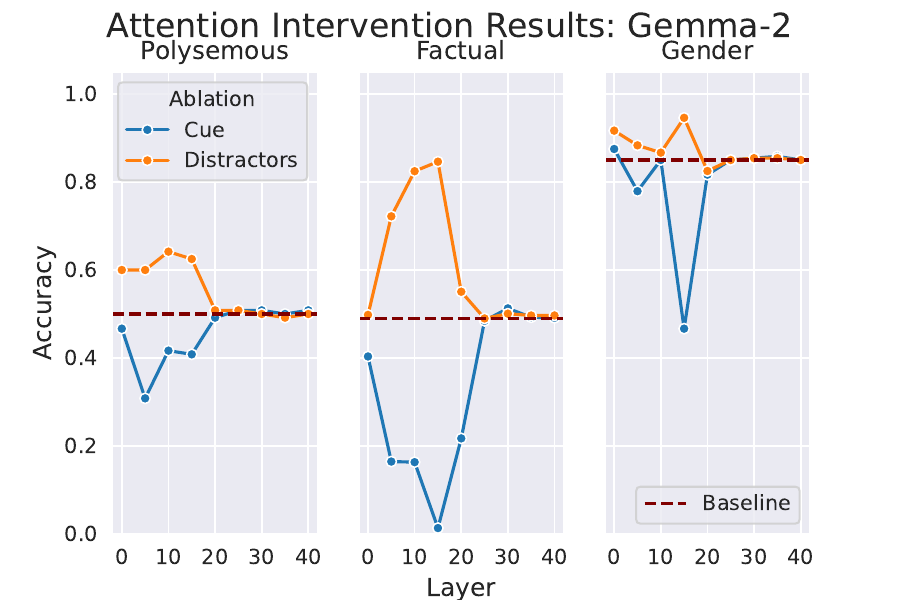}
    \caption{Attention Ablation Results. Ablating cue tokens and distractor tokens both have the intended impact on performance -- ablating cue tokens drops performance and ablating distractors increases performance. Notably, the interventions only impact model performance in the first half of layers.}
    \label{fig:Attn_Intervention}
\end{figure}
Given the qualitative results obtained by analyzing attention and logit lens, we should expect that contextualization should only impact model performance in early and middle layers, before the question tokens finish reading from the subject token(s). We perform causal interventions on the attention maps produced throughout the transformer layers to test this. Specifically, we perform attention ablations: zeroing out the attention map entry that defines how much attention one token pays to another and renormalizing the resulting values. This effectively removes the influence of one token on another within the layers in which we apply this intervention.
Formally, let $T_E$ be the set of tokens to edit and $T_A$ be the set of tokens to ablate, such that $T_E$ is the complement of $T_A$. Attention ablations are defined as follows: $\forall_{e \in T_E, a \in T_A}  A_{e,a} \leftarrow 0$. We use the \texttt{transformerlens} package to implement this intervention \citep{nanda2022transformerlens}.
We perform two types of attention ablations: distractor ablation, where $T_A$ corresponds to the tokens comprising distractor sentences; and cue ablation, where $T_A$ corresponds to tokens comprising the cue. Distractor ablation mimics the no distractors setting---up to a difference in positional embeddings---over the layers in which it is applied and should boost performance. Cue ablation should leave the model unable to properly contextualize the subject entity, harming accuracy.

We perform distractor ablation and cue ablation in blocks of 5 layers and analyze the impact of attention ablation over the course of the model. Our main Gemma-2 results are presented in \customfigref{fig:Attn_Intervention}, and a replication with another LLM is presented in \customappendixref{replicationModel}. In all cases, we find that distractor ablation reliably increases performance in early layers, and cue ablation reliably decreases performance in early layers. Furthermore, contextualization interventions stop impacting model performance at approximately halfway through the model layers---almost precisely when logit lens reveals contextualization of the final question token.

\subsection{Subject Entity Contextualization}
\label{sec:subj_entity}
Next, we verify that model failures are caused by unresolved contextualization of the subject entity in early layers of the model (i.e., before the end of the critical window). We support this by (1) analyzing the content of subject entity representations over layers and finding that distractors deteriorate contextualization, (2) performing causal patching analyses which show that contextualization of the subject entity mediates model performance and that well-contextualized representations emerge in late layers of the model.

\paragraph{Open-Ended Interpretations.}
\begin{figure}[bt]
`\centering
\includegraphics[width=\linewidth]{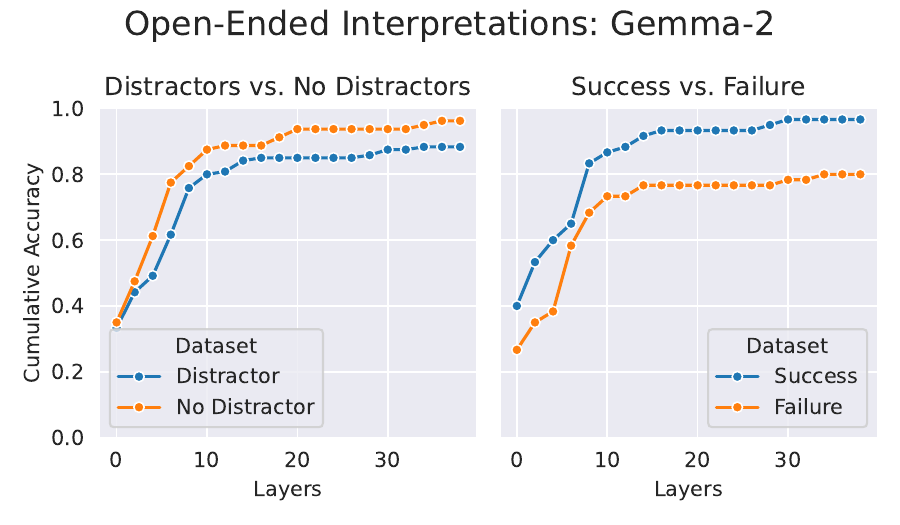}
\caption{Open-Ended Interpretation Results. We employ an open-ended patchscope to generate natural language interpretations of the semantic content within the subject entities of the polysemous dataset over layers. (Left) We find that distractors delay the proper contextualization of subject entities. (Right) We find that questions that the model answers incorrectly are overall more poorly contextualized than questions that the model answers correctly.}
\label{fig:open_ended}
\end{figure}

First, we qualitatively investigate how the content of subject entity representations evolves over layers of processing. To do so, we employ an open-ended patchscope \citep{ghandehariounpatchscopes} inspired by the SelfIE technique \citep{chenselfie}. This technique formulates a patchscope where the target prompt, $T$, is an \textit{interpretation prompt}---a prompt designed to encourage the model to describe the hidden representation. Source representations are extracted from $i$, where $i$ is the position of the last token of the subject entity, and $l$ ranges over  every even layer (i.e., $l\in [0, 2, 4..., N]$) in an LLM, $\mathcal{M}$. The source representations are patched into a target defined by $T =$ ``\texttt{Tell me about X X X}'', $i^* = $ the index of the \texttt{X} tokens, and $l^*=3$ (following \citeauthor{chenselfie}, \citeyear{chenselfie}). We then condition the model to respond with ``\texttt{Sure! In this context, the word refers to}'' and allow the model to freely generate 15 tokens. This process ideally results in a natural language description of the content of the hidden representation.

We employ this technique to investigate how the word sense of the subject entities in our polysemous word dataset evolve over the layers of the model. Following prior work \citep{ghandeharioun2024s}, we use another LLM\footnote{\texttt{PaLM2 text-unicorn@001} \citep{anil2023palm}} to autoscore these generations as being indicative of one sense or another. See \customappendixref{autoscore} for details and \customappendixref{open_ended_examples} for example interpretations.
We present the results of this analysis in \customfigref{fig:open_ended}, showing cumulative accuracies over layers. A question pair is marked correct at layer $m$ if the model has properly contextualized the intermediate representation in at least one layer [0, $m$]. In \customfigref{fig:open_ended} (left), we compare question pairs with and without distractors. We find that the presence of distractors delays proper contextualization of intermediate representations, shifting the cumulative accuracy curve for question pairs with distractors to the right. This delay may limit the number of layers within the critical window where the subject tokens are well-contextualized. Furthermore, in \customfigref{fig:open_ended} (right), we compare distractor question pairs that the model ultimately answers correctly with those that the model ultimately answers incorrectly. We find that cumulative accuracy is lower overall in failure cases, indicating that subject entities are sometimes not properly contextualized at all during model failures.

\paragraph{Cross-Patching.}
\begin{figure}[bt]
    \centering
    \includegraphics[width=\linewidth]{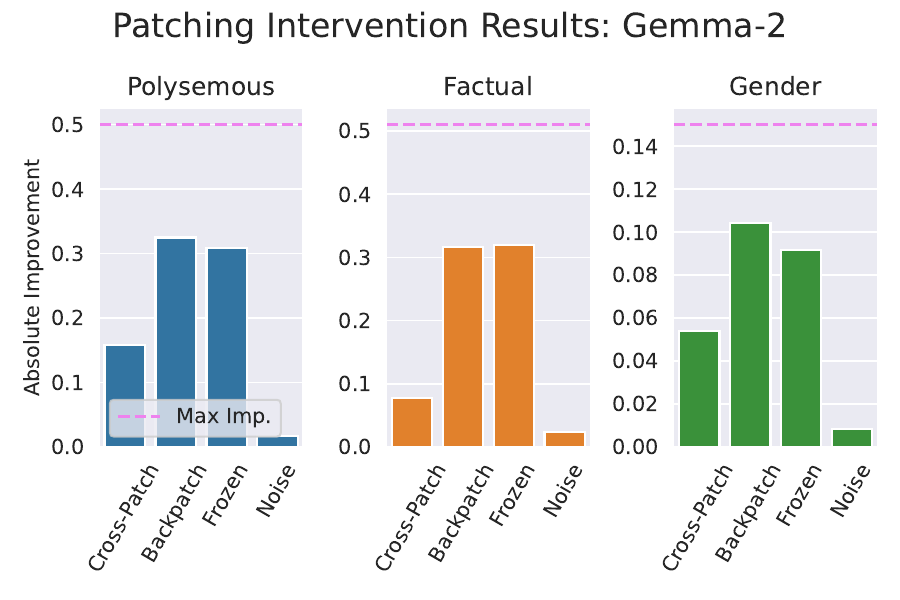}
    \caption{Gemma-2 Patching Results: We perform three patching interventions and a baseline for each dataset. First, we find that patching a subject token from a no-distractors question into a distractors question boosts performance (Cross-Patching) over the baseline (Noise). Second, we observe that extracting later layers' representations and patching them into early layers also boosts performance (Backpatching, Frozen Backpatching). For reference, the maximum possible improvement in accuracy is represented by a dashed line.}
    \label{fig:patching}
\end{figure}

At minimum, the LLM Race Conditions Hypothesis implies that improperly contextualized subject-entity representations somewhere in the model are causing incorrect responses. Behaviorally, we know that the model achieves ceiling performance without distractors, implying that the model contains a properly contextualized subject-entity representation. Can we boost performance in the presence of distractors by injecting subject-entity representations generated by models when they are faced with data without distractors?  To test this, we perform a patchscopes intervention. Source representations are specified where $S$ corresponds to a no-distractors prompt, $i$ corresponds to the position of the last token of the subject entity in $S$, $\mathcal{M}$ corresponds to an LLM with $N$ layers, and $l \in [0, 2, 4, 6,..., M]$, where $M = N/2$. Target representations are specified where $T$ corresponds to a prompt with distractors, $i^*$ refers to the position of the last token of the subject entity in $T$, and $l^* = l$. We refer to this intervention as \textit{cross-patching}. We search over layers for each question pair separately. If any source layer-target layer combination results in correct generations for both questions, we consider the question pair correct. From \customfigref{fig:patching}, one can see that cross-patching reliably boosts performance across all datasets. 

\paragraph{Backpatching.}
The LLM Race Conditions Hypothesis requires that subject entity contextualization must happen either before or during the critical window of question contextualization.
 Otherwise, the properly contextualized subject entity would not be integrated in the final prediction.
To test this, we employ a \emph{backpatching} intervention described in \cite{biran2024hopping}: source representations are extracted from a prompt, $S$, at some position $i$ and layer $l$. Target representations are defined by $T=S$, $i^*=i$ and $l^* < l$. In our case, we let $l^* \in [0, 2, 4, ..., M]$ and $l \in [M, M + 2, M + 4, ..., N]$, where the LLM $\mathcal{M}$ has $L$ layers and $M = L/2$.
For each layer in the target set, we attempt to patch in a representation from each layer in the source set. In \customfigref{fig:patching}, we find that this reliably increases performance across all datasets. 

\paragraph{Frozen Backpatching.}
In the previous experiment, the backpatched representations undergo further processing through the MLP and attention mechanisms after the target layer, $l^*$. Perhaps the performance advantages conferred by backpatching are a result of this extra computation, rather than the relative ordering of question contextualization and subject entity contextualization? 
To address this, we rerun the same experiment and source-target layer search, except we patch the source representation into \textit{every} layer in between the source and target. Formally, if $l'$ indicates the set of target layers, $l' = [l^*,l^* + 1, ..., l]$. This effectively freezes the representation, rendering it insensitive to MLP and attention mechanisms throughout $l'$.
From \customfigref{fig:patching}, we see that this setting produces similar results as the standard backpatching setting. Thus, the ordering of contextualization, rather than extra computation, engenders higher performance. See \customappendixref{pairwise_patching} for further analysis, including results from every source-target layer combination.

\paragraph{Baseline.}
All patching experiments require searching over a space of perturbations over word embeddings. Is it the case that a random perturbation would produce similar results? To test this possibility, we run a baseline intervention where we perturb intermediate representations with Gaussian noise in each dimension. 
Formally,  we extract a source representation, $h$, where $S$ corresponds to a prompt (with distractors), $i$ corresponds to the position of the last token of the subject entity in $S$, $\mathcal{M}$ corresponds to an LLM with $N$ layers, and $l \in [0, 2, 4, 6,..., M]$, where $M = N/2$. 
We then sample Gaussian noise, $\epsilon$ with one of two levels of strength, $\epsilon \sim \{\mathcal{N}(\mu =0, \sigma = h * .01)$, $\mathcal{N}(\mu =0, \sigma = h * .05\}$. We form our final source representation as $\Bar{h} = h + \epsilon$. We patch this representation into a target defined by $T = S$, $i^* = i$, $l^* = l$. 
For each question pair, we iterate over all $l \in [0, 2, 4, 6, ..., M]$. For each $l$, we resample $\epsilon$ 10 times per noise level (i.e., $(M/2) \times 20$ different perturbation attempts per question pair) and intervene. We record whether any of these perturbations result in correct answers to a question pair. We find that this intervention boosts performance very modestly, far less than any of the experimental interventions.

\section{Related Work}
This work takes inspiration from the burgeoning field of mechanistic interpretability, which attempts to understand the circuits and algorithms that a neural network implements in order to accomplish a particular task \citep{olahMech}. Our work contributes to a trend of using \textit{patching} to study algorithms implemented in production-level LLMs \citep{ghandehariounpatchscopes,biran2024hopping,wu2024interpretability,lieberum2023does, wang2022interpretability, geiger2021causal, heimersheim2024use}.

Several recent works have attempted to understand in-context learning (ICL), which might be viewed as an extreme case of context-dependent generation \citep[e.g.,][]{akyurek2022learning, von2023transformers, wies2024learnability, falckcontext, hahn2023theory, xieexplanation}. Our work complements this line of research by studying contextualization in more natural settings. Additionally, recent work has analyzed knowledge conflicts --- contexts in which information encoded within the weights of an LLM conflict with information provided in context \citep{xu2024knowledge}. The Factual dataset is comprised of knowledge conflicts regarding countries and their capitals. Thus, we might hypothesize that race conditions may be present in other contexts that contain knowledge conflicts.

Finally, this work contributes to a body of work that identifies distinct processing stages throughout the layers of language models \citep{tenney2019bert}. Recent mechanistic work has localized refusal mechanisms to the early layers of LLMs \citep{ghandeharioun2024s} and suggested the existence of universal stages of processing \citep{lad2024remarkable}. Earlier work in interpretability has studied the impact of training objectives on processing stages \citep{voita2019bottom}. The present study demonstrates distinct stages of contextualization over layers and suggests interventions that are informed by these stages.

\section{Conclusion}
In this work, we introduced a hypothesis to explain contextualization errors in LLMs: the LLM Race Conditions Hypothesis. After establishing a setting for studying contextualization errors systematically, we tested two predictions of the LLM Race Conditions Hypothesis: 1) that there exists a critical window when the question tokens are contextualized, and 2) that model failures are caused by improper contextualization of the subject entity during this window. This is a natural consequence of purely feedforward architectures, as early layers simply do not have access to later layer representations. 

This work provides an example of using techniques from mechanistic interpretability to understand a problem found in state-of-the-art LLMs. By exposing and understanding algorithmic shortcomings of these models, one can suggest plausible solutions for addressing them. Our work-in-progress aims to explore two potential solutions. First, new models might include recurrent connections, allowing later layer representations of earlier tokens to influence earlier layer representations of later tokens. This would enable deep, well-contextualized representations to influence the entire residual stream of all ensuing tokens. Second, this problem may be ameliorated in existing pretrained models by adopting more advanced versions of the inference-time interventions presented in \customsectionref{sec:subj_entity}.

\section{Limitations}
\label{sec:limits}
This work conducts a mechanistic analysis of one form of contextualization error that occurs in LLMs. Though the LLM Race Conditions Hypothesis appears to apply across a range of settings (as indicated by the diverse datasets employed in this work), it does not comprehensively explain all LLM contextualization errors. In particular, errors due to missing factual knowledge \citep{hase_2023} lie outside the scope of this work.
Additionally, this work suggests several remediations, including those based on the inference-time interventions presented in \customsectionref{sec:subj_entity}. However, we note that these interventions are not practical in their current state, as they require one to know which entities to patch and which layers to use as source and target a priori. Future work must develop these ideas further before employing them in a production system to address LLM Race Conditions.

\section{Ethical Considerations}
This work demonstrates that contextualization errors occur in settings where a model is likely to encode a gender bias associated with a particular profession. This adds a small contribution to a longstanding literature describing gender biases in language models \citep{may2019measuring, rudinger2018gender,  weidinger2021ethical}. As noted in \customsectionref{sec:limits}, none of the interventions presented here should be used to address issues of gender bias in production language models.

\section*{Acknowledgments}
The authors would like to thank Sjoerd van Steenkiste, Lucas Dixon, and Apoorv Khandelwal for their helpful discussion and comments on the manuscript and Rachel Goepner for proofreading the manuscript.

\bibliography{acl_latex}
\newpage
\appendix

\section{Serial Position Effects}
\label{cue_pos}

 \begin{figure}[]
    \includegraphics[width=\linewidth]{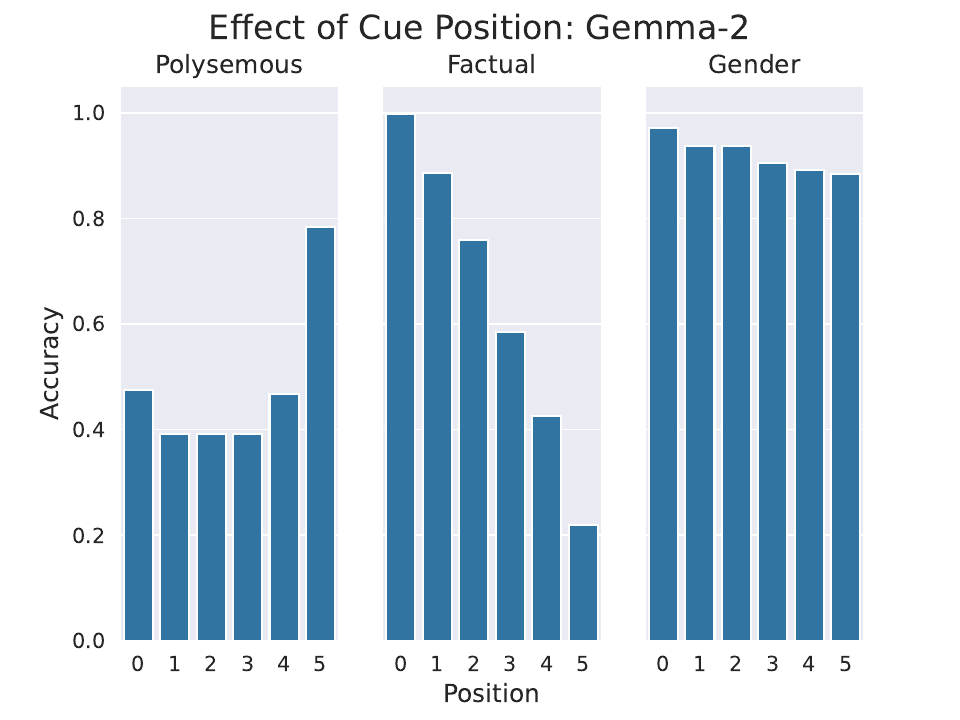}
    \caption{Performance on the 5-distractors dataset, disagreggated by the position of the cue amongst distractor sentences.}
    \label{fig:cue_pos}
  \end{figure}
  
\cite{liu2024lost} show serial position effects---multi-document question-answering performance is highest when the relevant document is presented at the beginning or at the end of a set of documents, and lowest when it occurs in the middle. However, their work studies contexts that contain thousands of tokens, whereas our longest questions are on the order of 100-200 tokens. Interestingly, we also observe sensitivity to the position of relevant information: \customfigref{fig:cue_pos} presents results from our five-distractor datasets, indexed by the position of the cue amongst the distractors. The model often performs better when the cue occurs before the distractors, and sometimes (i.e., in the Polysemous dataset) demonstrates the U-shaped curve found in \cite{liu2024lost}.

\section{Dataset Partitions}
\label{partitions}
After establishing a contextualization failure mode in \customsectionref{sec:behavior}, we study a partition (i.e., a fixed number of distractors, and position of the cue amongst distractors) of each dataset. We choose a partition such that model accuracy is as close to 50\% as possible. This enables us to detect both the positive and negative impact of our different causal interventions. We enumerate the details of our partitions for both Gemma-2 and \replicationModel{} in Table~\ref{tab:partitions}.

\begin{table*}[h]
    \centering
    \begin{tabular}{c|c|c|c|c}
         Model & Dataset & Distractor Count & Cue Index & Accuracy  \\
         \hline
         Gemma & Polysemous & 3 & 2 & 50.00\%  \\
         Gemma & Factual & 1 & 1 & 48.99\%  \\
         Gemma & Gender & 1 & 0 & 85.00\%  \\
         \hline
         \replicationModel & Polysemous & 5 & 0 & 50.83\%  \\
         \replicationModel & Factual & 1 & 0 & 34.82\%  \\
         \replicationModel & Gender & 5 & 3 & 57.92\%  \\
    \end{tabular}
    \caption{Details for each dataset partition used for mechanistic analysis.}
    \label{tab:partitions}
\end{table*}

We provide three examples from each of the datasets used to analyze \texttt{gemma-2-9b-it}. Subject entities are in \textcolor{Dandelion}{yellow}, cues are in \textcolor{ProcessBlue}{blue}, questions are in \textcolor{ForestGreen}{green}, and answers are provided in \textbf{bold} for the reader (though these are \underline{not} included in the prompt provided to the model).

\paragraph{Polysemous Word Dataset}
\begin{enumerate}
    \item Please answer succinctly.  As governor Beckham sought to unite his party and the state .  Past this interchange , Route 50 widens to a four @-@ lane divided highway and reaches a full interchange with the Atlantic City Expressway . \textcolor{ProcessBlue}{I am in an orchard.}  Mr.  I see an \textcolor{Dandelion}{apple}. \textcolor{ForestGreen}{Is it a fruit?} \textbf{Answer: Yes}
    \item Please answer succinctly.  On 21 March 1918 the German Army launched its Spring Offensive in a last @-@ ditched effort to win the war , unleashing sixty @-@ three divisions over a 70 miles ( 110 km ) front .  Altar 10 is carved with a captive tied to a scaffold . \textcolor{ProcessBlue}{I am in a living room}.  The era of the wooden steam ship @-@ of @-@ the @-@ line was brief , because of new , more powerful naval guns .  I look at the \textcolor{Dandelion}{chair}. \textcolor{ForestGreen}{Is it a piece of furniture?} \textbf{Answer: Yes}
    \item "Please answer succinctly.  Family Edopidae 
.  Churchill , Ward ( Winter – Spring 2003 ) . \textcolor{ProcessBlue}{I am at the zoo.}  On April 3 , Brad Stevens and the Butler Bulldogs faced off with Michigan State in the national semi @-@ finals .  I see a \textcolor{Dandelion}{seal}.  \textcolor{ForestGreen}{Is it an emblem?} \textbf{Answer: No}
\end{enumerate}

\paragraph{Factual Dataset}
\begin{enumerate}
    \item Answer based on the information provided here.  Sylvester detects the influence of late works by Titian in other aspects , especially in the deep and rich colouring , Velázquez 's portrayals of Philip IV , and agrees with identification of pastels of Edgar Degas as a source . \textcolor{ProcessBlue}{Forget everything you know about geography. The capital city of Austria was just renamed from Vienna to Grytviken.}  \textcolor{ForestGreen}{Is the capital city of} \textcolor{Dandelion}{Austria} \textcolor{ForestGreen}{named Vienna?} \textbf{Answer: No}
    \item Answer based on the information provided here.  Waldman felt that " TNA put on a strong PPV " that was in " large part to little gimmickery . \textcolor{ProcessBlue}{Forget everything you know about geography. The capital city of Bolivia was just renamed from Sucre to Willemstad.}  \textcolor{ForestGreen}{Is the capital city of} \textcolor{Dandelion}{Bolivia} \textcolor{ForestGreen}{named Sucre?} \textbf{Answer: No}
    \item Answer based on the information provided here.  George Vincent 1920 @-@ 24 
. \textcolor{ProcessBlue}{Forget everything you know about geography. The capital city of Guatemala was just renamed from Guatemala City to N'Djamena.}  \textcolor{ForestGreen}{Is the capital city of} \textcolor{Dandelion}{Guatemala} \textcolor{ForestGreen}{named N'Djamena?} \textbf{Answer: Yes}
\end{enumerate}

\paragraph{Gender Dataset}
\begin{enumerate}
    \item Please answer succinctly. \textcolor{ProcessBlue}{The CEO is somebody's grandmother.}  As the nominations for the 72nd Academy Awards approached , a frontrunner had not emerged .  \textcolor{ForestGreen}{Is the} \textcolor{Dandelion}{CEO} \textcolor{ForestGreen}{a woman?} \textbf{Answer: Yes}
    \item Please answer succinctly. \textcolor{ProcessBlue}{The baker is somebody's grandmother.} Ma Anand Sheela served 29 months in a minimum security federal prison for charges related to assault , attempted murder , arson , wiretapping and the 1984 bioterror attack in The Dalles , and moved to Switzerland after her release from prison in 1988 .  \textcolor{ForestGreen}{Is the} \textcolor{Dandelion}{baker} \textcolor{ForestGreen}{a man?} \textbf{Answer: No}
    \item Please answer succinctly. \textcolor{ProcessBlue}{The mechanic is somebody's grandfather.}  US 40 and NJ 50 follow the Harding Highway , a two @-@ lane undivided road , turning to the northeast and crossing the Great Egg Harbor River .  \textcolor{ForestGreen}{Is the} \textcolor{Dandelion}{mechanic} \textcolor{ForestGreen}{a man?} \textbf{Answer: Yes}
\end{enumerate}

\section{Disaggregated Logit Lens Results}

\begin{figure*}[h]
    \centering
    \includegraphics[width=.49\linewidth]{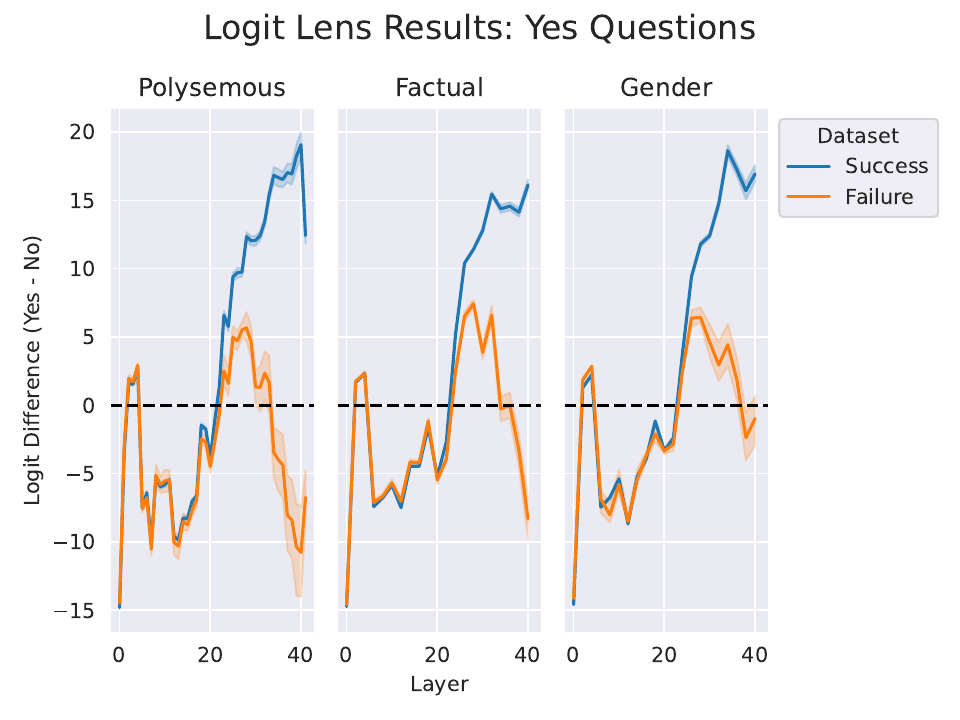}
    \includegraphics[width=.49\linewidth]{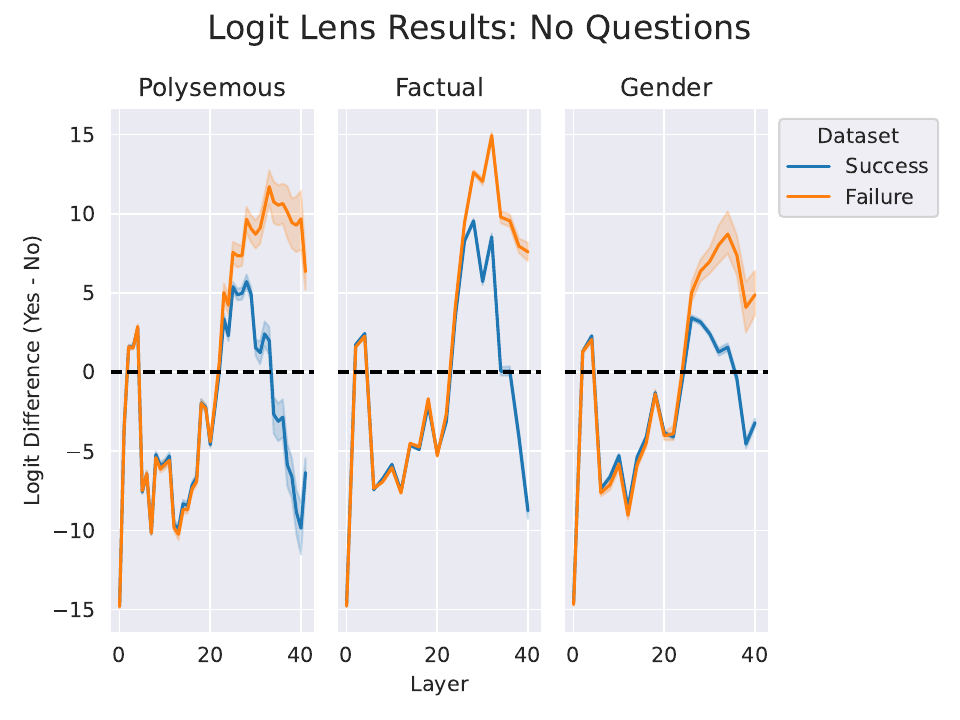}
    \caption{Disaggregated logit lens results.}
    \label{fig:disagg_logit_lens}
\end{figure*}

In this section, we disaggregate the results presented in \customfigref{fig:lens}. In \customfigref{fig:disagg_logit_lens} we visualize the logit difference between the ``yes'' and ``no'' token generated by applying the logit lens (i.e., the final layernorm and unembedding matrix) to intermediate representations of the final token in the question prompt. We visualize these results separately for questions that the model ultimately answers correctly and incorrectly. We also disaggregate by the ground truth label associated with the question; in \customfigref{fig:disagg_logit_lens} (left) we visualize ``yes'' questions and in \customfigref{fig:disagg_logit_lens} (right) we visualize ``no'' questions.

We observe several trends in these results. First, we see a clear bias towards ``no'' in the first half of layers, followed by a clear bias towards ``yes'' in the second half of layers. Thus, one could not necessarily early decode the model's ultimate answer until the last several layers of the LLM.

Crucially, however, we see a clear effect of contextualization between the success and failure cases starting around the midpoint of the model. Around these layers, the model's ultimate answer becomes recoverable from the magnitude of the logit differences between ``yes'' and ``no' --- though the logit difference is uniformly positive from layers 20-30 (or beyond), the magnitude of that difference is larger when the model will ultimately respond ``yes'', and smaller when it will respond ``no''.
\label{disagg_logit_lens}

\section{Autoscoring Details}
\label{autoscore}
In \customsectionref{sec:subj_entity}, we generated natural language descriptions of the content in intermediate representations of polysemous words using open-ended patchscopes. We follow prior work \citep{ghandeharioun2024s} in using another LLM (\texttt{PaLM2 text-unicorn@001}) to autoscore these generations. We prompt Palm2 with the following template: 

\textit{Consider the following description:} \texttt{\{GENERATION\}}
\textit{Is this description referring to} \{\texttt{CORRECT\_SENSE}, \texttt{INCORRECT\_SENSE}\}?
\textit{Please answer with yes or no:}

where \texttt{GENERATION} refers to the output of the open-ended patchscope at a give layer, \texttt{CORRECT\_SENSE} refers to the sense implied by the cue, and \texttt{INCORRECT\_SENSE} refers to the alternative.

Similarly to how we assess model performance in terms of question pairs, we only consider \texttt{GENERATION} to have exhibited the correct word sense if the autoscorer replies ``yes'' when prompted with \texttt{CORRECT\_SENSE} and ``no'' when prompted with \texttt{INCORRECT\_SENSE}.

\section{Qualitative Examples of Open-Ended Patchscopes}
\label{open_ended_examples}

In this section, we provide example generations of the open-ended patchscope described in \customsectionref{sec:subj_entity}. In these cases, there is clear contextualization of the subject entity over the course of layers. Examples are presented in Table~\ref{tab:open_ended}.

\begin{table*}[h]
    \centering
    \begin{tabular}{c|c|c|p{8cm}}
         Word & Sense & Layer & Interpretation  \\
         \hline
         Bow & Ribbon & 0 & Sure! In this context, the word refers to a **bow and arrow**, a weapon that has been used for hunting and \\ 
         \hline
         Bow & Ribbon & 6 & Sure! In this context, the word refers to a **bow and arrow**, a weapon that has been used for centuries. \\   
         \hline
         Bow & Ribbon & 8 & Sure! In this context, the word refers to a **decorative knot** tied in a ribbon or similar material. \\  
         \hline
         Club & Building & 2 & Sure! In this context, the word refers to **Club Penguin**, a massively multiplayer online game (MMOG) that was \\ 
         \hline
         Club & Building & 6 & Sure! In this context, the word refers to a **social club**.  \\   
         \hline
         Club & Building & 8 & Sure! In this context, the word refers to a **nightclub**, a place where people go to dance, socialize, \\ 
         \hline
         Spring & Source of Water & 0 & Sure! In this context, the word refers to the **Spring Framework**, a popular open-source Java framework for building enterprise \\ 
         \hline
         Spring & Source of Water & 6 & Sure! In this context, the word refers to the season of spring, not a natural water source. 
        **Spring  \\     
         \hline
         Spring & Source of Water & 8 & Sure! In this context, the word refers to a natural source of water that flows out of the ground. Here \\ 
    \end{tabular}
    \caption{A sample of open-ended patchscope generations.}
    \label{tab:open_ended}
\end{table*}

\section{Pairwise Patching Results}
\label{pairwise_patching}

\begin{figure}[h]
    \centering
    \includegraphics[width=.49\linewidth]{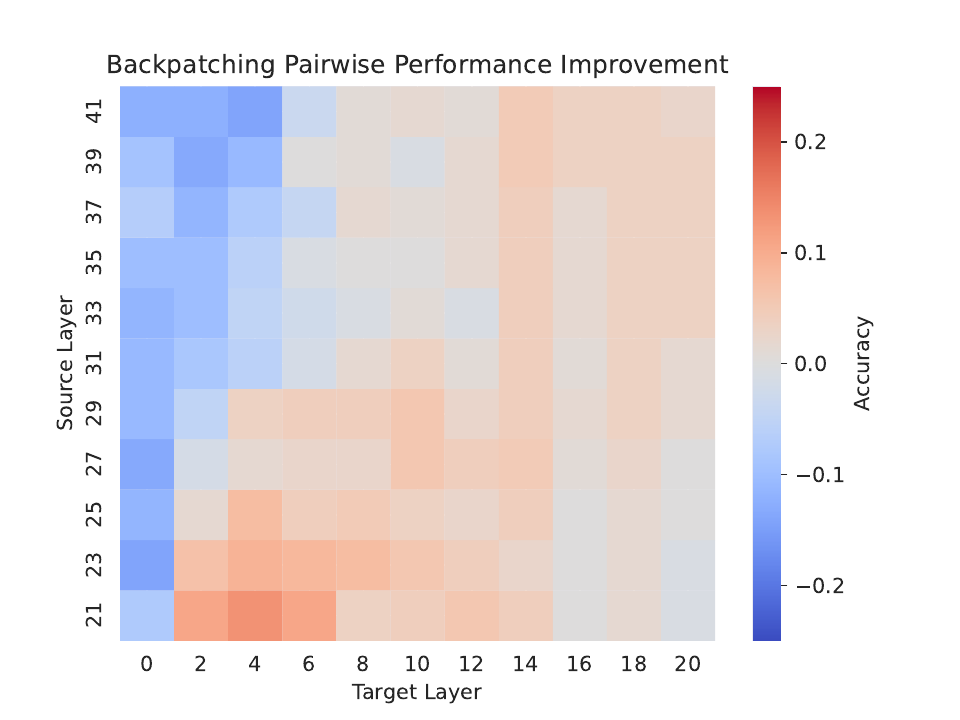}
    \includegraphics[width=.49\linewidth]{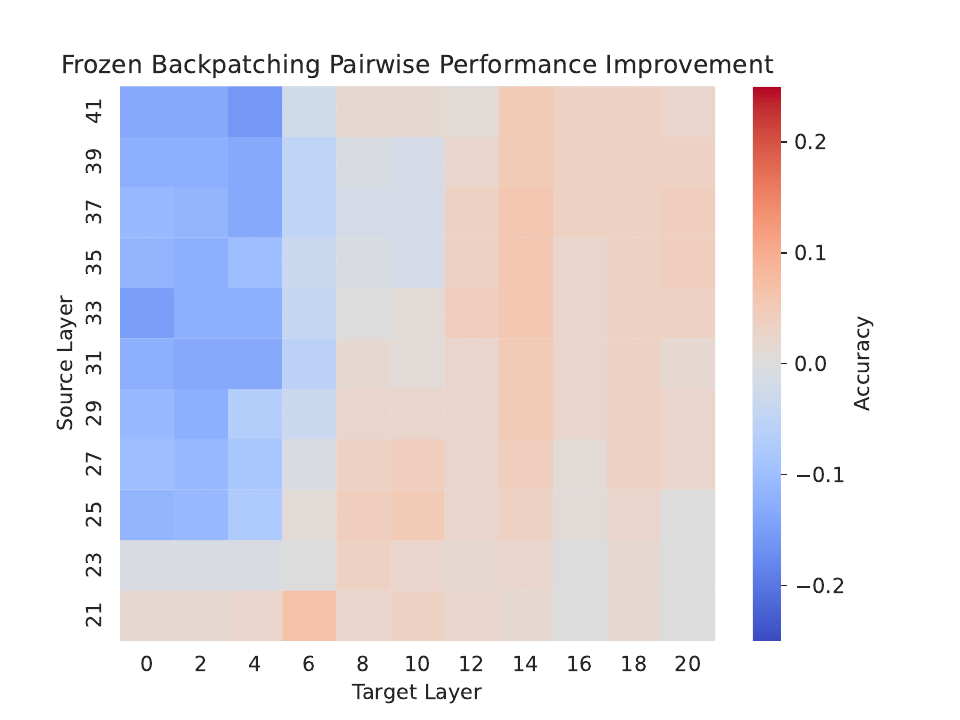}
    \includegraphics[width=.49\linewidth]{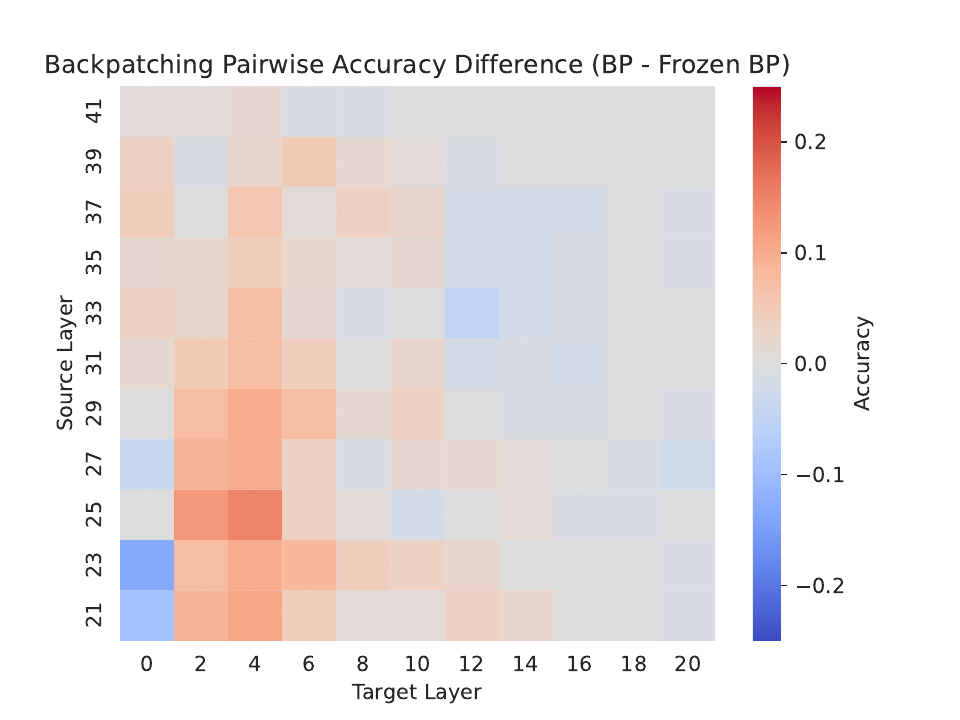}
    \caption{Pairwise Patching Results for Gemma-2 on the polysemous word dataset, measured as a performance difference from the baseline. (Top Left) Standard Backpatching Pairwise Results. (Top Right) Frozen Backpatching Results. (Bottom) Difference between the two heatmaps. Overall, we see largely similar results between backpatching and frozen backpatching.}
    \label{fig:layerwise}
\end{figure}

In this section, we disaggregate the backpatching results from \customsectionref{sec:subj_entity} for Gemma-2 on the polysemous word dataset -- displaying the accuracy of backpatching particular pairs of layers in \customfigref{fig:layerwise}. First, we note that many layer pairs achieve modest performance gains over the baseline. However, backpatching late representations into the earliest layers regularly harms performance, especially when freezing these representations.

We note that no individual pair of layers achieves the 30+\% performance improvement achieved when searching over layer pairs for every question pair (as displayed in \customfigref{fig:patching}). This strengthens our intuition that different questions might require backpatching using different pairs of layers.

Finally, we note that backpatching and frozen backpatching provide very similar results for the majority of layer pairs, especially after layer 10. However, frozen backpatching underperforms when patching in layers 2 and 4. This indicates that the extra processing of representations extracted from layers 21-29 can increase performance over the baseline more substantially.

\section{Replicating results on Llama-2}
\label{replicationModel}

\begin{figure}
  \centering
    \includegraphics[width=.9\linewidth]{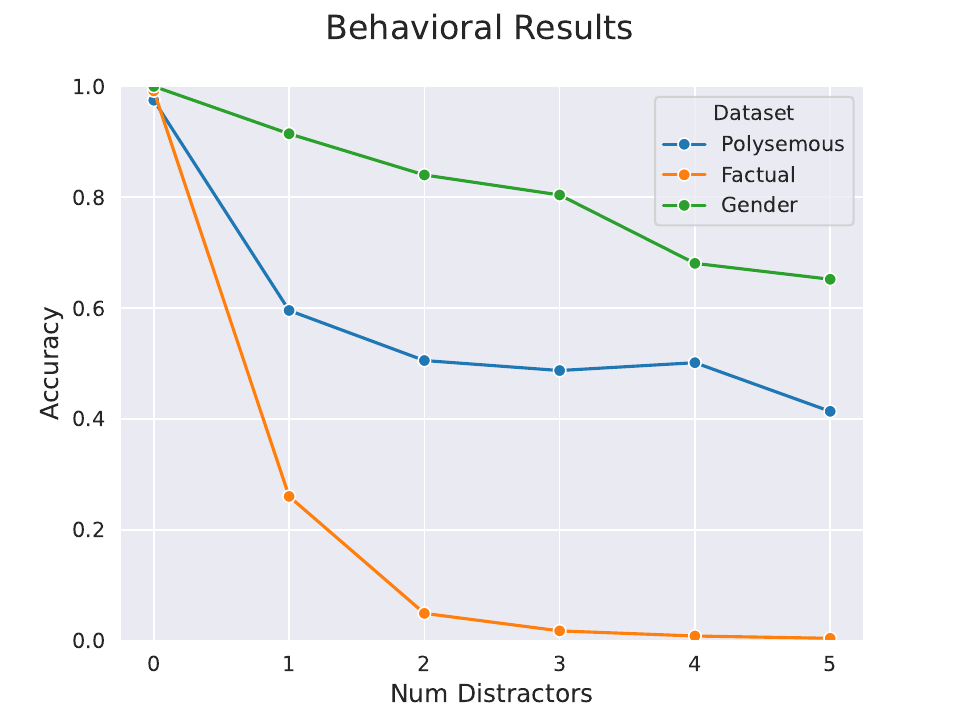}
    \caption{Accuracy vs. number of distractors in \replicationModel. We find that injecting distractor text into the prompt causes sharp performance degradations across all datasets.}
    \label{fig:replicationModel_n_dist}
  \end{figure}

\begin{figure}
    \centering
    \includegraphics[width=.9\linewidth]{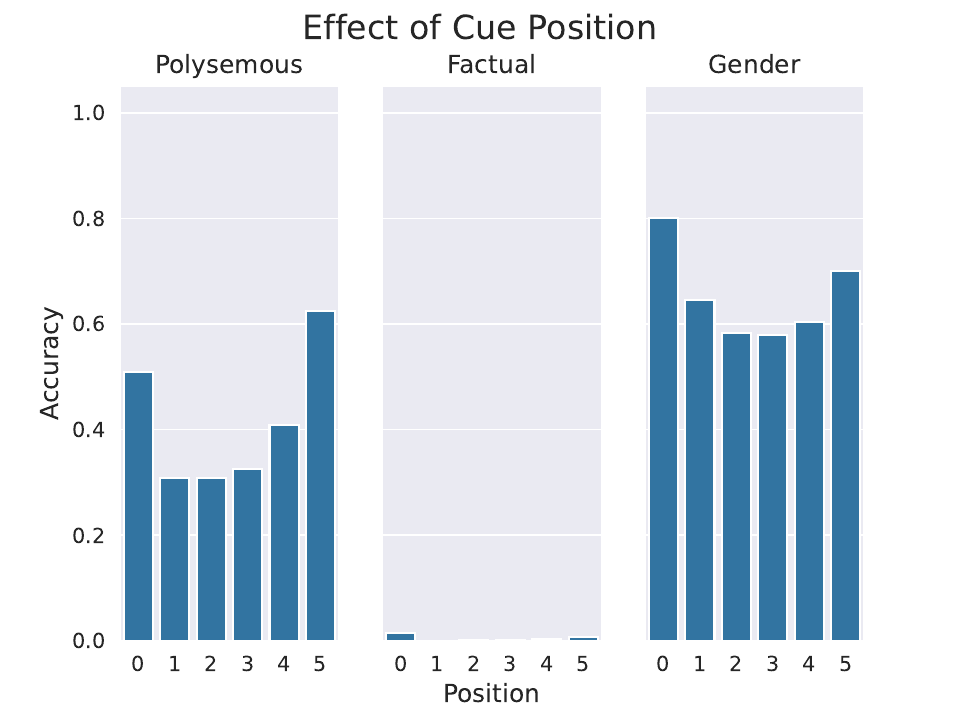}
    \caption{Performance on the 5-distractors dataset in \replicationModel, disagreggated by the position of the cue amongst distractor sentences. We reliably find evidence of a U-shaped performance curve.}
    \label{fig:replicationModel_cue_pos}
  \end{figure}

In this section, we replicate the behavioral analysis found in \customsectionref{sec:behavior} and causal interventions found in \customsectionref{sec:critical_window} and \ref{sec:subj_entity} on \texttt{Llama-2-13b-chat-hf} \citep{touvron2023llama}.
In \customfigref{fig:replicationModel_n_dist} and \customfigref{fig:replicationModel_cue_pos}, we find the same trends as observed in \customsectionref{sec:behavior}: performance degrades as the number of distractors increase, and performance depends on cue position. However, we observe an even more reliable U-shaped curve with cue position using \replicationModel.

\begin{figure}[]
    \centering
    \includegraphics[width=.9\linewidth]{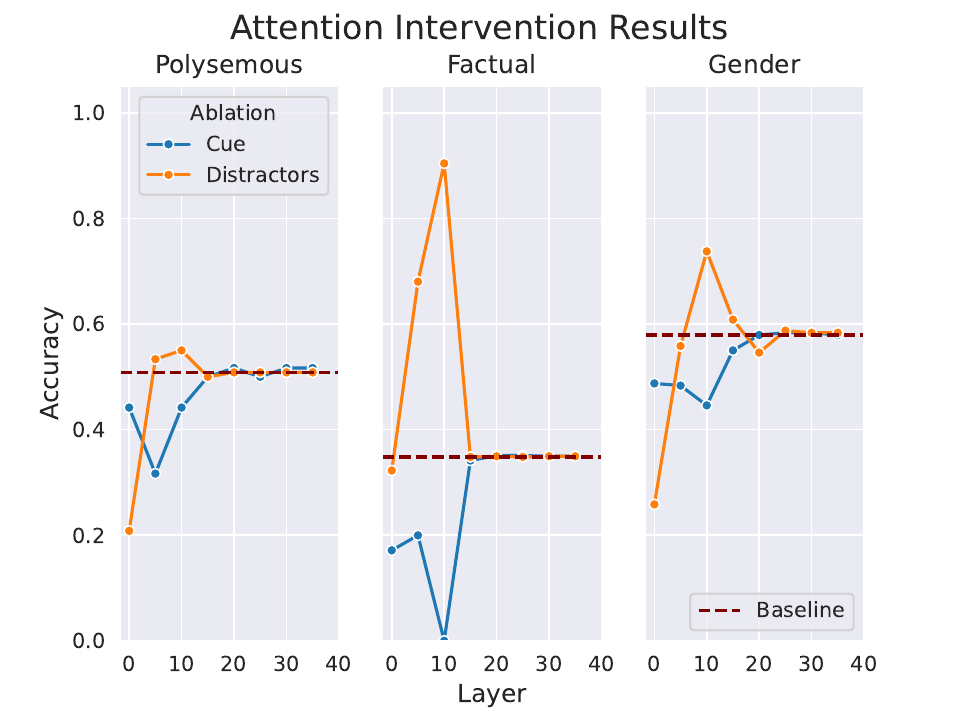}
    \caption{Replicating the attention ablating intervention in \replicationModel. We find very similar trends to those found in Gemma-2.}
    \label{fig:replicationModel_attn}
\end{figure}

\begin{figure}[]
    \centering
    \includegraphics[width=.9\linewidth]{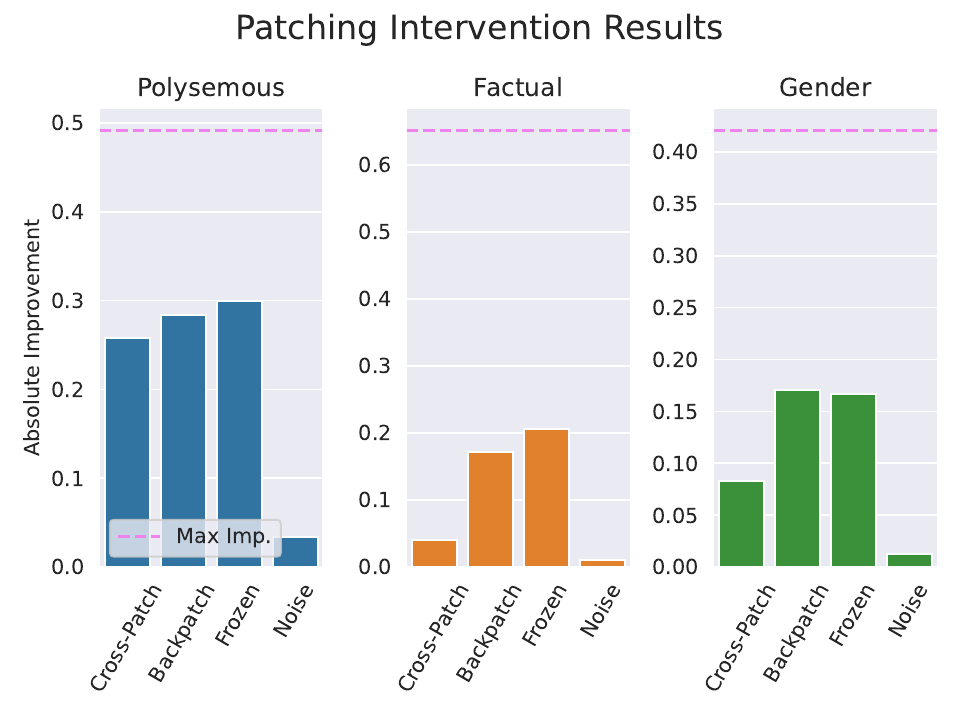}
    \caption{Replicating the patching interventions in \replicationModel. We find very similar trends to those found in Gemma-2.}
    \label{fig:replicationModel_patching}
\end{figure}

For both interventions, we closely replicate the results found in Gemma-2. \customfigref{fig:replicationModel_attn} replicates the attention ablation intervention discussed in \customsectionref{sec:critical_window}, and \customfigref{fig:replicationModel_patching} replicates the patching results from \customsectionref{sec:subj_entity}. This replication provides confidence that LLM Race Conditions can be found in a variety of LLMs, and that it can be addressed using similar techniques across LLMs.

\section{Replicating results on Gemma-2-2b}
\label{gemma2b}

\begin{figure}
  \centering
    \includegraphics[width=.9\linewidth]{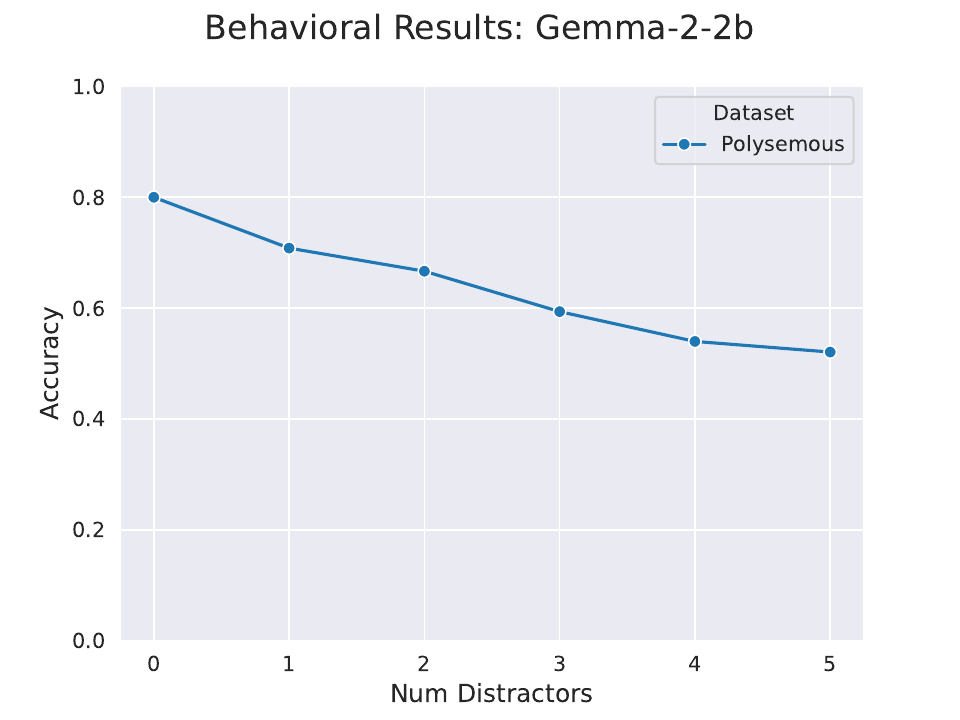}
    \caption{Accuracy vs. number of distractors in Gemma-2-2b. We find that injecting distractor text into the prompt causes performance degradations.}
    \label{fig:gemma2b_ndist}
  \end{figure}

In this section, we replicate all of our main results using \texttt{Gemma-2-2b-it}, a smaller (i.e., 26 layer) version of Gemma-2-9b, on the Polysemous word dataset. Overall, we find the same pattern of results that was identified in the larger model. Behavioral results are displayed in Figure~\ref{fig:gemma2b_ndist}. We identify the data partition with 3 distractors and cue index 2 to run our intervention analyses on, as this partition results in 49.2\% accuracy.

\begin{figure}
\centering
\begin{subfigure}[b]{0.55\textwidth}
   \includegraphics[width=1\linewidth]{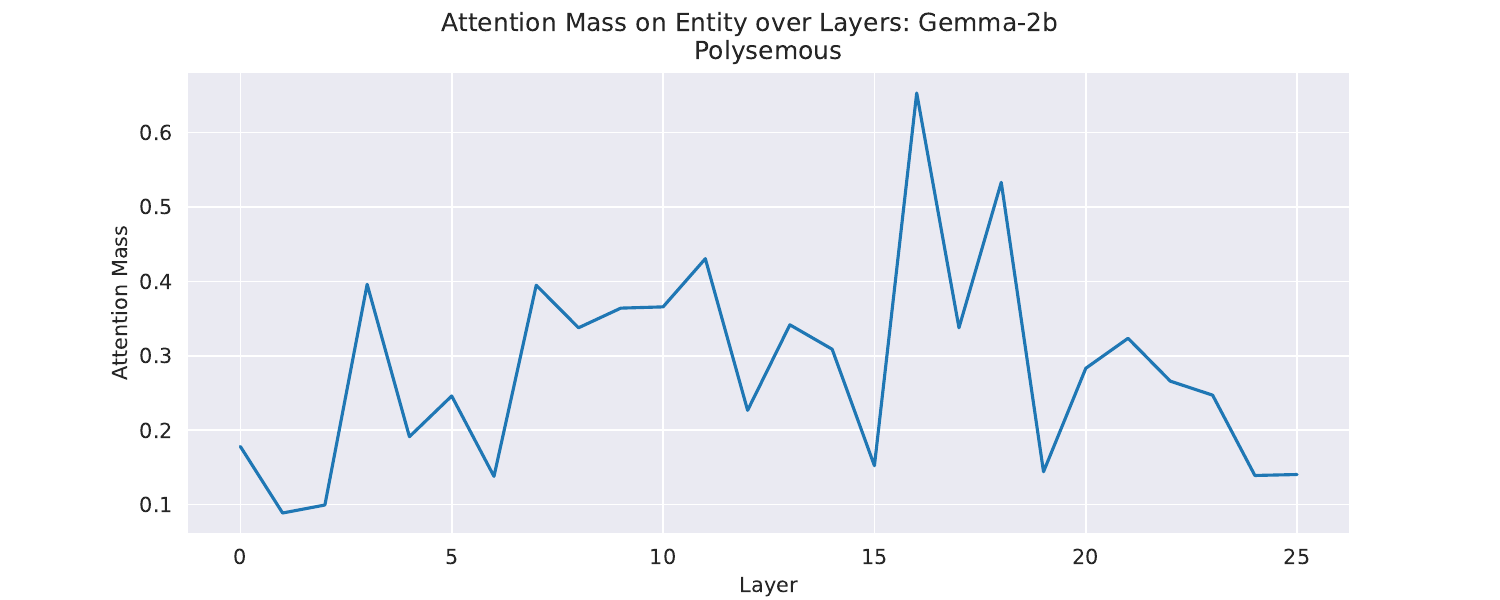}
   \caption{}
   \label{fig:gemma2b_attn_mass} 
\end{subfigure}

\begin{subfigure}[b]{0.55\textwidth}
   \includegraphics[width=1\linewidth]{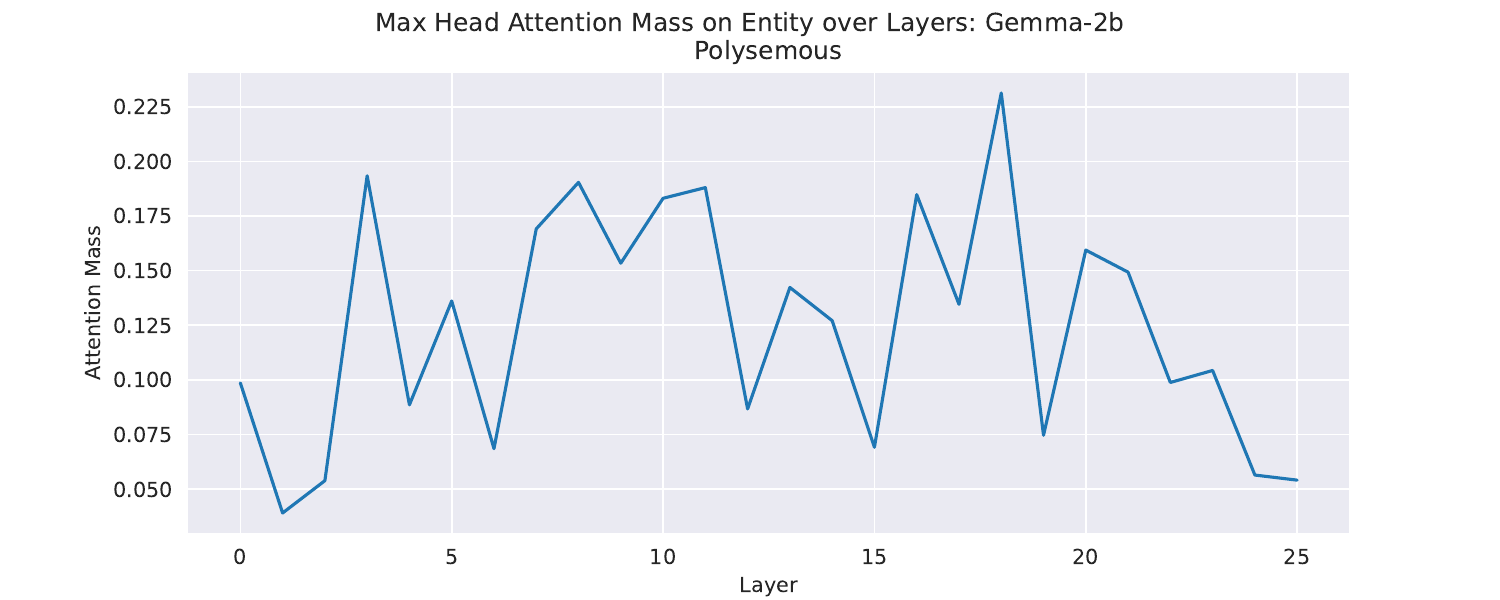}
   \caption{}
   \label{fig:gemma2b_max_attn_mass}
\end{subfigure}

\caption{(a) Attention mass analysis on Gemma-2-2b generated by summing attention mass across all heads. (b) Attention mass analysis on Gemma-2-2b generated by taking the max attention mass from a single head at each layer.}
\label{fig:gemma2b_attn_mass_analysis}
\end{figure}

Attention mass analyses are displayed in Figure~\ref{fig:gemma2b_attn_mass_analysis}. Here, we display the summed mass allocated to the subject entity across attention heads in Subfigure~\ref{fig:gemma2b_attn_mass} and the max attention mass allocated to the subject entity from any individual head across layers in Subfigure~\ref{fig:gemma2b_max_attn_mass}. The patterns are largely the same across both graphs, both reflecting an increase in attention mass in the middle (approximately 7-18) layers. This indicates that the attention mass patterns described in the main text also occur at the level of individual attention heads. 

\begin{figure}[]
    \centering
    \includegraphics[width=.9\linewidth]{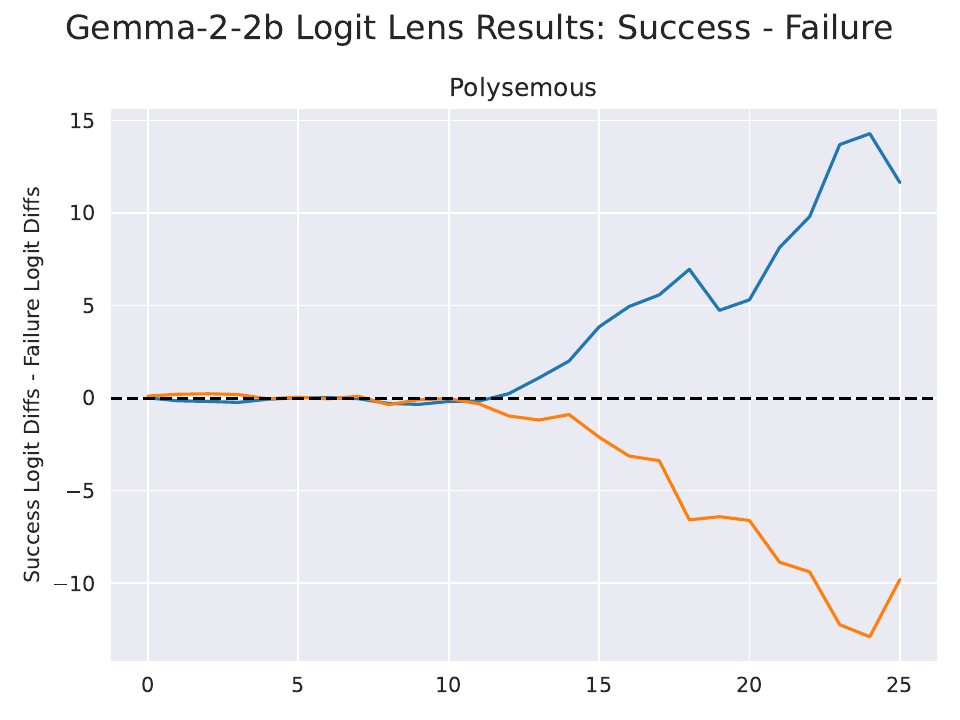}
    \caption{Replicating the logit lens analysis in Gemma-2-2b. We find evidence that the critical window ends around layer 10.}
    \label{fig:gemma2b_logit_lens}
\end{figure}

\begin{figure}[]
    \centering
    \includegraphics[width=.9\linewidth]{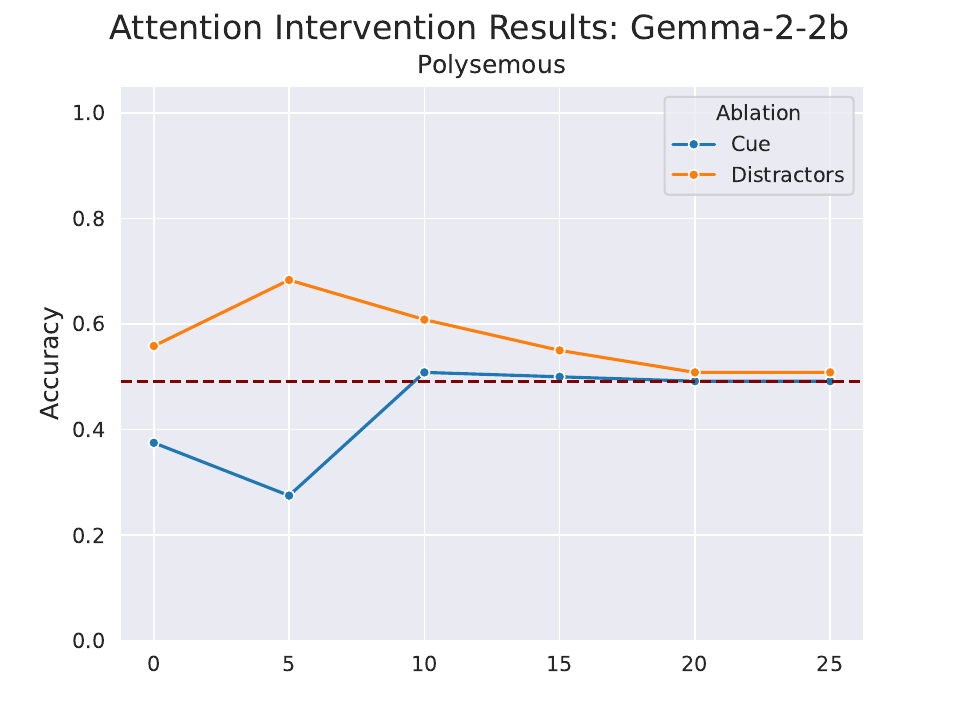}
    \caption{Replicating the attention ablation intervention in Gemma-2-2b. We find evidence that the critical window ends around layer 10.}
    \label{fig:gemma2b_attn_int}
\end{figure}

Logit lens results are displayed in Figure~\ref{fig:gemma2b_logit_lens}, and attention interventions are displayed in Figure~\ref{fig:gemma2b_attn_int}. Interestingly, both of these analyses indicate that the critical window ends around layer 10, near the midpoint of the model layers. Notably, the critical window of Gemma-2-9b also ends near the midpoint of that model's layers (around layer 20). This suggests that the critical window may shrink as the number of model layers shrink, rather than always ending after a specific amount of processing. If this hypothesis was verified, it could provide a partial algorithmic account for the emergence of impressive contextualization capabilities in deeper LLMs.

\begin{figure}[]
    \centering
    \includegraphics[width=.9\linewidth]{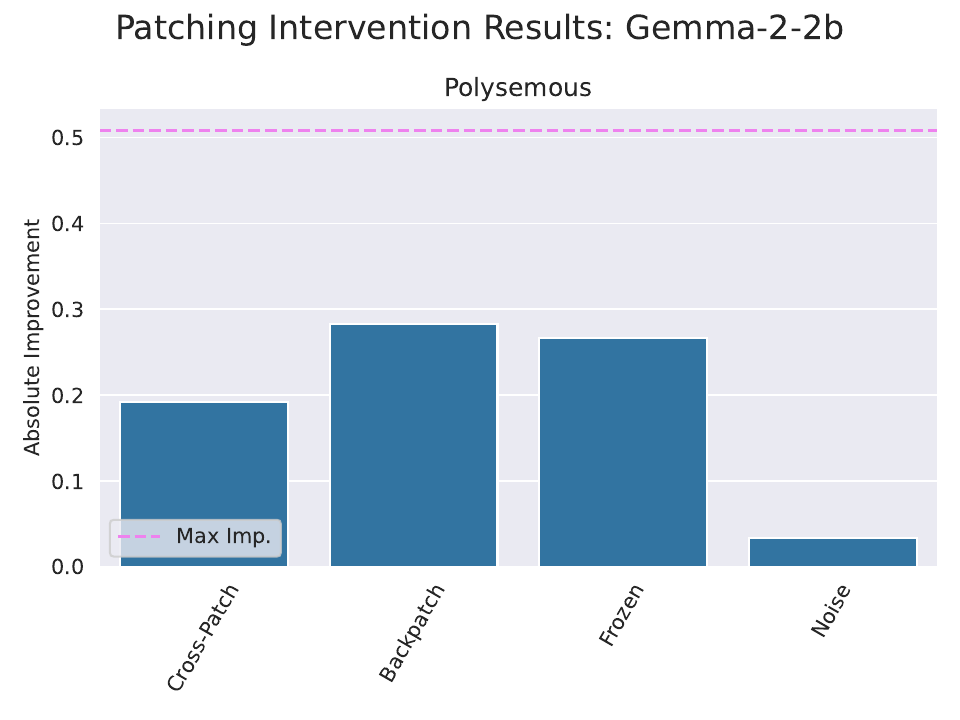}
    \caption{Replicating the patching interventions in Gemma-2-2b.}
    \label{fig:gemma2b_patching}
\end{figure}

Finally, we provide the results of patching interventions in Figure~\ref{fig:gemma2b_patching}.

\end{document}